\documentclass[11pt,a4paper]{article}
\usepackage[hyperref]{acl2021}
\usepackage{times}
\usepackage{latexsym}
\usepackage{graphicx}
\usepackage{enumitem}
\usepackage{booktabs}
\usepackage{multirow}
\usepackage{amssymb}

\newcommand\rurl[1]{%
  \href{https://#1}{\nolinkurl{#1}}%
}
\usepackage{microtype}

\aclfinalcopy 

\title{Scientific Credibility of Machine Translation Research: \\
A Meta-Evaluation of 769 Papers}

\author{Benjamin Marie\qquad Atsushi Fujita  \qquad Raphael Rubino  \\
  National Institute of Information and Communications Technology \\
	3-5 Hikaridai, Seika-cho, Soraku-gun, Kyoto, 619-0289, Japan \\
  {\{bmarie,atsushi.fujita,raphael.rubino\}@nict.go.jp}}

\date{}

\begin{document}
\maketitle
\begin{abstract}
This paper presents the first large-scale meta-evaluation of machine translation (MT). We annotated MT evaluations conducted in 769 research papers published from 2010 to 2020. Our study shows that practices for automatic MT evaluation have dramatically changed during the past decade and follow concerning trends. An increasing number of MT evaluations exclusively rely on differences between BLEU scores to draw conclusions, without performing any kind of statistical significance testing nor human evaluation, while at least 108 metrics claiming to be better than BLEU have been proposed. MT evaluations in recent papers tend to copy and compare automatic metric scores from previous work to claim the superiority of a method or an algorithm without confirming neither exactly the same training, validating, and testing data have been used nor the metric scores are comparable. Furthermore, tools for reporting standardized metric scores are still far from being widely adopted by the MT community. After showing how the accumulation of these pitfalls leads to dubious evaluation, we propose a guideline to encourage better automatic MT evaluation along with a simple meta-evaluation scoring method to assess its credibility. 
\end{abstract}

\section{Introduction}
New research publications in machine translation (MT) regularly introduce new methods and algorithms to improve the translation quality of MT systems. In the literature, translation quality is usually evaluated with automatic metrics such as BLEU \citep{papineni-etal-2002-bleu} and, more rarely, by humans. To assess whether an MT system performs better than another MT system, their scores given by an automatic metric are directly compared. While such comparisons between MT systems are exhibited in the large majority of MT papers, there are no well-defined guideline nor clear prerequisites under which a comparison between MT systems is considered valid. Consequently, we assume that evaluation in MT is conducted with different degrees of thoroughness across papers and that evaluation practices have evolved over the years. What could be considered, by the research community, as a good evaluation methodology ten years ago may not be considered good today, and vice versa. This evolution has not been studied and whether MT evaluation has become better, or worse, is debatable.

On the other hand, several requirements for MT evaluation have been well-identified. For instance, the limitations of BLEU are well-known \citep{callison-burch-etal-2006-evaluating,reiter-2018-structured,mathur-etal-2020-tangled} and the necessity to report automatic metric scores through standardized tools, such as SacreBLEU, has been recognized \citep{post-2018-call}. Moreover, a trustworthy evaluation may adopt statistical significance testing \citep{koehn-2004-statistical} and strong baselines \citep{denkowski-neubig-2017-stronger}. However, to what extent these requirements have been met in MT publications is unclear.

In this paper, we propose the first large-scale meta-evaluation of MT in which we manually annotated 769 research papers published from 2010 to 2020.
Our study shows that evaluation in MT has dramatically changed since 2010. An increasing number of publications exclusively rely on BLEU scores to draw their conclusions. The large majority of publications do not perform statistical significance testing, especially since 2016. Moreover, an increasing number of papers copy and compare BLEU scores published by previous work while tools to report standardized metric scores are still far from being extensively adopted by the MT community. We also show that compared systems are often trained, validated, or even evaluated, on data that are not exactly the same.
After demonstrating how the accumulation of these pitfalls leads to dubious evaluation, we propose a general guideline for automatic evaluation in MT and a simple scoring method to meta-evaluate an MT paper. We believe that the adoption of these tools by authors or reviewers have the potential to reverse the concerning trends observed in this meta-evaluation.

\section{A Survey on MT Evaluation}
We manually annotated the MT evaluation in research papers published from 2010 to 2020 at *ACL conferences.\footnote{We considered only *ACL main conferences, namely ACL, NAACL, EACL, EMNLP, CoNLL, and AACL, as they are the primary venues for publishing MT papers.}
To identify MT papers, we searched the ACL Anthology website\footnote{\rurl{www.aclweb.org/anthology/}} for the terms ``MT'' or ``translation'' in their titles\footnote{There are potentially MT papers falling outside these search criteria but we considered the 769 papers we obtained to be representative enough for the purpose of this study.} and analyzed among them the 769 papers that make comparisons of translation quality between at least two MT systems. For each year between 2010 and 2020, we respectively annotated the following numbers of papers: 53, 40, 59, 80, 67, 45, 51, 62, 94, 115, and 103.

We annotated each paper as follows:
\begin{enumerate}[label=A\arabic*.]
    \item All the automatic metrics used to evaluate the translation quality of MT systems. We did not list variants of the same metric: e.g., chrF3 and chrF++ are labeled chrF \cite{popovic-2015-chrf}. Moreover, we did not consider metrics which only target specific aspects of the translation quality, such as pronoun translation and rare word translation.
    \item Whether a human evaluation of the translation quality has been conducted: yes or no. If the human evaluation only targets specific types of errors and did not evaluate the translation quality of the entire text, we answered ``no.''\footnote{Note that we only check here whether the automatic evaluation is supported by a human evaluation. Previous work already studied pitfalls in human evaluation \citep{laubli2020set}.}
    \item Whether any kind of statistical significance testing of the difference between automatic metric scores has been performed: yes or no. Potentially, some papers did perform significance testing without mentioning it, but due to the lack of evidences such papers have been annotated with ``no'' for this question.
    \item Whether it makes comparisons with automatic metric scores directly copied from previous work to support its conclusion: yes or no. Most papers copying scores (mostly BLEU) clearly mention it. If there is no evidence that the scores have been copied, we annotated these papers with ``no'' for this question.
    \item Whether SacreBLEU has been used: yes or no. If there is no mention or reference to ``SacreBLEU,'' we assume that it has not been used. Note that ``yes'' does not mean that the paper used SacreBLEU for all the MT systems evaluated.
    \item If previous work has not been reproduced but copied, whether it has been confirmed that all the compared MT systems used exactly the same pre-processed training, validating, and testing data: yes or no. 
\end{enumerate}
Except for A6, the annotation was straightforward since most papers present a dedicated section for experimental settings with most of the information we searched for. Answering A6 required to check the data exploited in the previous work used for comparison.
Note that answering ``yes'' to the questions from A2 to A6 may only be true for at least one of the comparisons between MT systems, while we did not evaluate how well it applies. For instance, answering ``yes'' to A5 only means that at least one of the systems has been evaluated with SacreBLEU but not that the SacreBLEU signature has been reported nor that SacreBLEU scores have been correctly compared with other BLEU scores also computed with SacreBLEU.

Our annotations are available as a supplemental material of this paper.
To keep track of the evolution of MT evaluation, we will periodically
update the annotations and will make it available online.\footnote{The up-to-date version can be found here: \rurl{github.com/benjamin-marie/meta_evaluation_mt}.}

\begin{figure*}[t]
   \centering
   \includegraphics[scale=0.51]{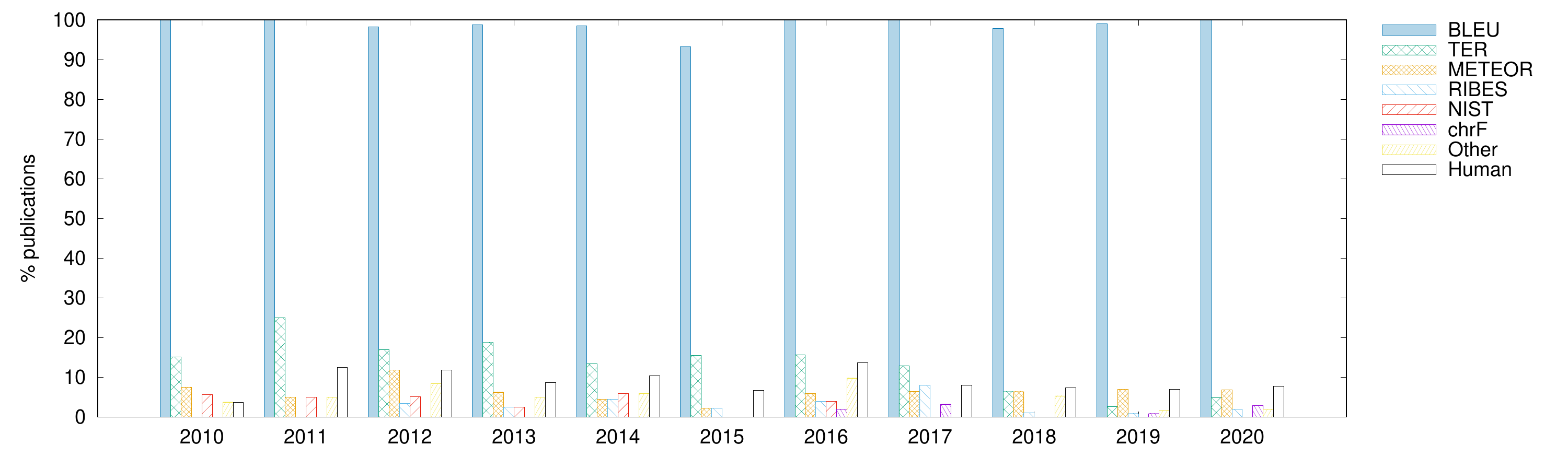}
   \caption{\label{fig:metricpaper} Percentage of papers using each evaluation metric per year. Metrics displayed are used in more than five papers. ``Other'' denotes all other automatic metrics. ``Human'' denotes that a human evaluation has been conducted.}
\end{figure*}

\section{Pitfalls and Concerning Trends}
\label{sec:pitfalls}
This section discusses the four main pitfalls identified in our meta-evaluation of MT: the exclusive use of BLEU, the absence of statistical significance testing, the comparison of incomparable results from previous work, and the reliance on comparison between MT systems that do not exploit exactly the same data.
We report on how often they affected MT papers and recent trends. Based on previous work and supporting experiments, we show how each of these problems and their accumulation lead to scientifically dubious MT evaluation.
\subsection{The 99\% BLEU}
\label{sec:bleu}
Automatic metrics for evaluating translation quality have numerous advantages over a human evaluation. They are very fast and virtually free to run provided that a reference translation is already available. Their scores are also reproducible. As such, automatic metrics remained at the center of MT evaluation for the past two decades. New metrics that better correlate with human judgments are regularly introduced. We propose in this section to analyze the use of automatic metrics in MT research, relying on our annotations for A1 and A2.

This is probably the most expected finding in our study: the overwhelming majority of MT publications uses BLEU. Precisely, 98.8\% of the annotated papers report on BLEU scores. As shown in Figure~\ref{fig:metricpaper}, the ratio of papers using BLEU remained stable over the years.
On the other hand, BLEU scores used to be more often supported by scores from other metrics, such as TER \citep{Snover06astudy} and METEOR \citep{banerjee-lavie-2005-meteor}, than they are now. The large majority of papers, 74.3\%, only used BLEU scores to evaluate MT systems, i.e., without the support of any other metrics nor human evaluation. It increases to 82.1\% if we consider only the years 2019 and 2020. 

This tendency looks surprising considering that no less than 108 new metrics\footnote{We did not count variants of the same metric and excluded metrics only proposed for an evaluation at segment level.} have been proposed in the last decade. They have been shown to better correlate with human judgments than BLEU. Some are even easier to use and more reproducible by being tokenization agnostic, such as chrF. We counted 29 metrics proposed at *ACL conferences since 2010 while the remaining metrics were proposed at the WMT Metrics Shared Tasks. 89\% of these 108 new metrics have never been used in an *ACL publication on MT (except in the papers proposing the metrics). Among these metrics, only RIBES \citep{isozaki-etal-2010-automatic} and chrF have been used in more than two MT research paper.

\begin{table*}[t]
    \centering
    \scriptsize
    \begin{tabular}{c|lc|lc|lc|lc}
    \toprule
        \multirow{2}{*}{Rank} &  \multicolumn{4}{c|}{Japanse-to-English (Ja$\rightarrow$En)} & \multicolumn{4}{c}{Chinese-to-English (Zh$\rightarrow$En)}\\  
         & BLEU & System & chrF & System & BLEU & System & chrF & System \\
         \midrule
1 &26.6$^\spadesuit$&NiuTrans& 0.536&Tohoku-AIP-NTT&36.9&WeChat\_AI&0.653&Volctrans\\
2 &25.5&Tohoku-AIP-NTT& 0.535&NiuTrans&36.8&Tencent\_Translation&0.648$^\blacklozenge$&Tencent\_Translation\\
3 &24.8$^\blacklozenge$&OPPO& 0.523$^\blacklozenge$&OPPO&36.6&DiDi\_NLP&0.645$^\blacklozenge$&DiDi\_NLP\\
4 &22.8$^\blacklozenge$&NICT\_Kyoto& 0.507$^\blacklozenge$&Online-A&36.6&Volctrans&0.644$^\blacklozenge$&DeepMind\\
5 &22.2$^\blacklozenge$&eTranslation& 0.504$^\blacklozenge$&Online-B&35.9$^\blacklozenge$&THUNLP&0.643$^\blacklozenge$&THUNLP\\
    \bottomrule     
    \end{tabular}
    \caption{Rankings of WMT20 top 5 submissions for the News Translation Shared Tasks according to BLEU and chrF scores. 
    Superscripts indicate systems that are significantly worse ($^\blacklozenge$) and better ($^\spadesuit$) according to each metric ($p$-value $<$ 0.05) than Tohoku-AIP-NTT and Volctrans systems for Ja→En and Zh→En, respectively.
    }
    \label{tab:wmtchrfbelu}
\end{table*}

When properly used, BLEU is a valid metric for evaluating translation quality of MT systems \citep{callison-burch-etal-2006-evaluating, reiter-2018-structured}. Nonetheless, we argue that better metrics proposed by the research community should be used to improve MT evaluation. To illustrate how wrong an evaluation can become by only relying on one metric, we computed with BLEU and chrF scores\footnote{SacreBLEU (short) signatures: chrF2+l.\{ja-en,zh-en\}+n.6+s.false+t.wmt20+v.1.5.0 and BLEU+c.mixed+l.\{ja-en,zh-en\}+\#.1+s.exp+t.wmt20+tok.13a+v.1.5.0} of WMT20 submissions to the news translation shared task\footnote{\rurl{data.statmt.org/wmt20/translation-task/}} \citep{barrault-etal-2020-findings} using SacreBLEU and show rankings given by both metrics in Table~\ref{tab:wmtchrfbelu}. Results show that BLEU and chrF produce two different rankings. For instance, for the Ja$\rightarrow$En task, NiuTrans system is the best according to BLEU by being 1.1 points better than the Tohoku-AIP-NTT system ranked second. In most MT papers, such a difference in BLEU points would be considered as a \emph{significant} evidence of the superiority of an MT system and as an improvement in translation quality. Relying only on these BLEU scores without any statistical significance testing nor human evaluation would thus lead to the conclusion that NiuTrans system is the best. However, according to another metric that better correlates with human judgment, i.e., chrF, this does not hold: Tohoku-AIP-NTT system is better. Similar observations are made for the Zh$\rightarrow$En task.\footnote{For both Ja$\rightarrow$En and Zh$\rightarrow$En tasks, systems ranked first by chrF were also ranked first by the human evaluation.} These observations have often been made by the MT community, for instance at WMT shared tasks, but nonetheless rarely seen in research papers.

We assume that MT researchers largely ignore new metrics in their research papers for the sake of some comparability with previous work or simply because differences between BLEU scores may seem more meaningful or easier to interpret than differences between scores of a rarely used metric. Most papers even qualify differences between BLEU scores as ``small,'' ``large,'' or ``significant'' (not necessarily statistically), implying that there is a scientific consensus on the meaning of differences between BLEU scores. As we show in the following sections, all these considerations are illusory. Moreover, BLEU may also be directly requested by reviewers, or even worse, other metrics may be requested to be dropped.\footnote{Examples of such requests or related comments by reviewers can be found in the ACL 2017 review corpus (\rurl{github.com/allenai/PeerRead}), e.g., in the review ID 369 we read: ``I am also rather suspicious of the fact that the authors present only METEOR results and no BLEU.''} We believe that the exclusive reliance on BLEU can be ended and the use of better metrics should be encouraged, in addition to or in lieu of BLEU, by the adoption of a guideline for automatic MT evaluation (see Section~\ref{sec:guideline}).

\subsection{The Disappearing Statistical Significance Testing}
Statistical significance testing is a standard methodology designed to ensure that experimental results are not coincidental. In MT, statistical significance testing has been used on automatic metric scores and more particularly to assess whether a particular difference of metric scores between two MT systems is not coincidental. Two methods are prevalent in MT: the paired bootstrap test \citep{koehn-2004-statistical} and the approximate randomization test \citep{riezler-maxwell-2005-pitfalls}, for instance respectively implemented in Moses\footnote{\rurl{github.com/moses-smt/mosesdecoder}} and MultEval.\footnote{\rurl{github.com/jhclark/multeval}}

\begin{figure}[t]
   \centering
   \includegraphics[scale=0.51]{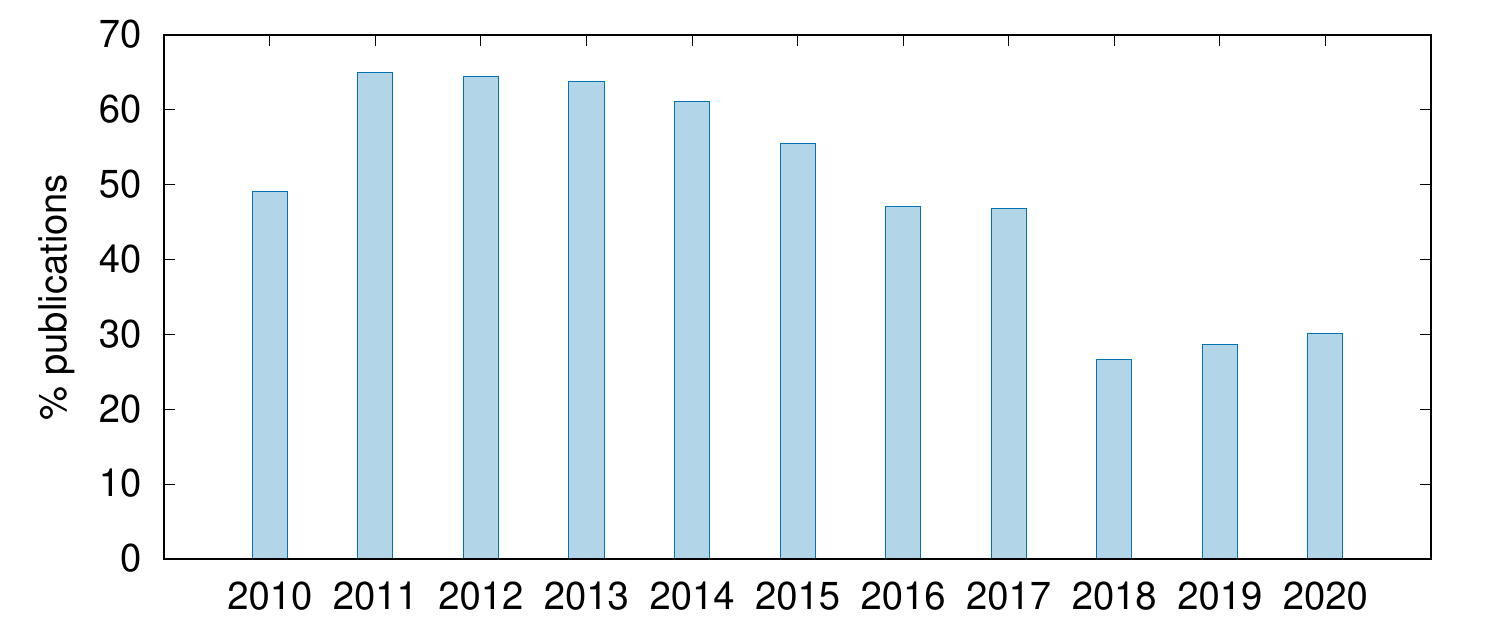}
   \caption{\label{fig:sigpaper} Percentage of papers testing statistical significance of differences between metric scores.}
\end{figure}

\citet{dror-etal-2018-hitchhikers} report that while the natural language processing (NLP) community assigns a great value to experimental results, statistical significance testing is rarely used. We verified if this applies to MT evaluations based on our annotations for A3. Figure~\ref{fig:sigpaper} shows the percentage of papers that performed statistical significance testing. We found out that the observations by \citet{dror-etal-2018-hitchhikers} apply to MT since never more than 65.0\% of the publications in a year (2011) performed statistical significance testing. Furthermore, our meta-evaluation shows a sharp decrease of its use since 2016. Most papers did not check whether their results are not coincidental but drew conclusions from them. 

MT papers mainly relied on the amplitude of the differences between metric scores to state whether they are significant or not. This was also observed by \citet{dror-etal-2018-hitchhikers} for NLP in general.

For illustration, we also performed statistical significance testing\footnote{For all the statistical significance testing performed in this paper, we used the paired bootstrap test with 1,000 samples and 1,000 iterations.} with BLEU and chrF scores on the WMT20 submissions in Table~\ref{tab:wmtchrfbelu}. For Ja$\rightarrow$En, NiuTrans system is significantly better in BLEU than Tohoku-AIP-NTT system. In contrast, they are not significantly different in chrF. Using only BLEU, we would conclude that NiuTrans system is significantly the best. This is not confirmed by chrF hence we need to report on more than one metric score to conduct a credible evaluation, even when performing statistical significance testing.

Furthermore, to show that the significance of a difference between metric scores is independent from its amplitude, we performed additional experiments by modifying only one sentence, replacing it with an empty line or by the repetition of the same token many times,\footnote{This could be considered as a simulation of potential defects from an MT framework or model, e.g., when translating extremely long sequences.} from Tohoku-AIP-NTT and Volctrans systems' outputs. Results in BLEU and chrF are reported in Table~\ref{tab:wmtcustom}. We observe that a difference in only one sentence can lead to a difference in BLEU of 6.8 points (Ja$\rightarrow$En, Custom 2).\footnote{For ``Custom 2,'' BLEU greatly penalized the increase of the number of tokens in the output. This is indicated by the length ratio reported by SacreBLEU but rarely shown in MT papers.} Nonetheless, our statistical significance tests did not find any system significantly better than the others.

\begin{table}[t]
    \centering
    \scriptsize
    \begin{tabular}{ccc|ccc}
    \toprule
          \multicolumn{3}{c|}{Ja$\rightarrow$En} & \multicolumn{3}{c}{Zh$\rightarrow$En}\\  
         System & BLEU & chrF & System & BLEU & chrF \\
         \midrule
Tohoku-AIP-NTT & 25.5 & 0.536 &Volctrans & 36.6 & 0.653 \\
\midrule
Custom 1 & 25.5 & 0.536 & Custom 1 & 36.6 & 0.653 \\
Custom 2 & 18.7 & 0.503 & Custom 2 & 32.2 & 0.638 \\

    \bottomrule     
    \end{tabular}
    \caption{BLEU and chrF scores of the customized Tohoku-AIP-NTT and Volctrans outputs from which only one sentence has been modified. The first row shows the results of the original WMT20 submissions. Custom 1 replaced the last sentence with an empty line, while Custom 2 replaced the last sentence with a sequence repeating 10k times the same token. None of these systems are significantly different according to statistical significance testing on these scores.}
    \label{tab:wmtcustom}
\end{table}

While the importance of statistical significance testing is regularly debated by the scientific community \citep{doi:10.1080/00031305.2019.1583913}, it remains one of the most cost-effective tools to check how trustworthy a particular difference between two metric scores is.\footnote{\citet{doi:10.1080/00031305.2019.1583913} give several recommendations for a better use of statistical significance testing.}

\subsection{The Copied Results}
\begin{figure}[t]
   \centering
   \includegraphics[scale=0.51]{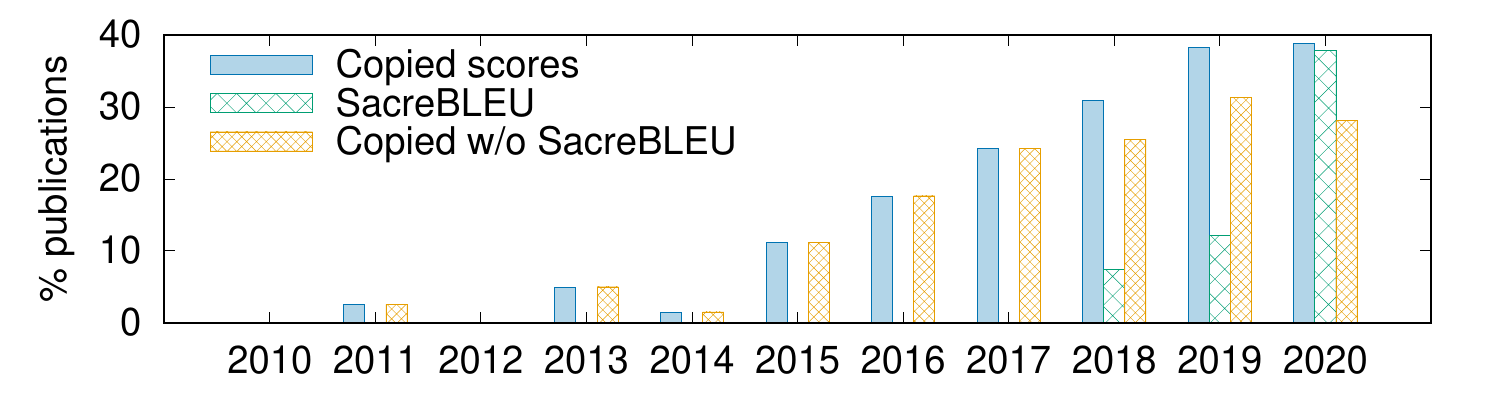}
   \caption{\label{fig:sbcppaper} Percentage of papers copying scores from previous work (``Copied scores''), using SacreBLEU (``SacreBLEU''), and copying scores without using SacreBLEU (``Copied w/o SacreBLEU'').}
\end{figure}

An MT paper may compare the automatic metric scores of proposed MT systems with the scores reported in previous work. This practice has the advantage to save the time and cost of reproducing competing methods.
Based on our annotations for A4, we counted how often papers copied the scores from previous work to compare them with their own scores. As pointed out by Figure~\ref{fig:sbcppaper}, copying scores (mostly BLEU) from previous work was rarely done before 2015. In 2019 and 2020, nearly 40\% of the papers reported on comparisons with scores from other papers.
While many papers copied and compared metric scores across papers, it is often unclear whether they are actually comparable. As demonstrated by \citet{post-2018-call}, BLEU, as for most metrics, is not a single metric. It requires several parameters and is dependent on the pre-processing of the MT output and reference translation used for scoring. In fact, \citet{post-2018-call} pointed out that most papers do not provide enough information to enable the comparability of their scores with other work. \citet{post-2018-call} proposed a tool, SacreBLEU, to standardize metrics\footnote{Currently BLEU, chrF, and TER.} in order to guarantee this comparability, provided that all the scores compared are computed with SacreBLEU.\footnote{SacreBLEU also generates a ``signature'' to further ensure this comparability: two scores computed through SacreBLEU with an identical signature are comparable.} This is the only tool of this kind used by the papers we annotated. However, based on our annotations for A5, Figure~\ref{fig:sbcppaper} shows that SacreBLEU is still far from widely adopted by the MT community, even though it is gradually getting more popular since its emergence in 2018. Moreover, papers that copy BLEU scores do not always use SacreBLEU, even in 2020. 

\begin{table}[t]
    \centering
    \scriptsize
    \begin{tabular}{l|cc|cc}
    \toprule
        \multirow{2}{*}{Processing} & \multicolumn{2}{c|}{Tohoku-AIP-NTT (Ja$\rightarrow$En)} & \multicolumn{2}{c}{Volctrans (Zh$\rightarrow$En)} \\
         & BLEU & chrF & BLEU & chrF \\
         \midrule
    original            & 25.5 & 0.536 & 36.6 & 0.653 \\
    fully lowercased    & 26.9 & 0.549 & 38.2 & 0.664 \\
    norm. punct.        & 25.5 & 0.537 & 37.8 & 0.657 \\
    \midrule
     tokenized          & 26.7 & 0.541 & 37.1 & 0.653 \\
     + norm. punct.     & 26.8 & 0.541 & 38.5 & 0.659 \\
     + aggressive       & 27.8 & 0.541 & 39.5 & 0.659 \\
    
    \bottomrule     
    \end{tabular}
    \caption{BLEU and chrF scores computed by SacreBLEU after applying different processing on some WMT20 MT system outputs (from Tohoku-AIP-NTT and Volctrans) and on the reference translations. None of these rows are comparable.}
    \label{tab:diffpreprocess}
\end{table}

To illustrate how deceiving a comparison of copied scores can be, we report on BLEU and chrF scores using different processing,\footnote{For all our processing, we used Moses (code version mmt-mvp-v0.12.1-2851-gc054501) scripts.} commonly adopted by MT researchers, applied to some MT system outputs and reference translations of the WMT20 news translation shared tasks. Our results are presented in Table~\ref{tab:diffpreprocess}. The first row presents original SacreBLEU scores, i.e., detokenized. Second and third rows respectively show the impact of lowercasing and punctuation normalization on metric scores. Scores are increased. Last three rows show the results on tokenized MT outputs. Applying both punctuation normalization and aggressive tokenization with Moses scripts leads to BLEU scores several points higher than the original SacreBLEU scores. Obviously, none of the scores in different rows are comparable. Nonetheless, MT papers still often report on \emph{tokenized} BLEU scores compared with tokenized, or even detokenized, BLEU scores from other papers without exactly knowing how tokenization has been performed. Tokenized BLEU scores reported in MT papers are often computed using the multi-bleu script of Moses even though it displays the following warning:\footnote{This warning has been added on 20 Oct. 2017. Insightful discussions on this commit can be found there:\\ \rurl{github.com/moses-smt/mosesdecoder/commit/545eee7e75487aeaf45a8b077c57e189e50b2c2e}.} ``\emph{The scores depend on your tokenizer, which is unlikely to be reproducible from your paper or consistent across research groups.}''

Even though the work of \citet{post-2018-call} is a well-acclaimed initiative towards better MT evaluation, we believe that it can only be a patch for questionable evaluation practices. A comparison with a copied score is \emph{de facto} associated with the absence of statistical significance testing since the MT output used to compute the copied score is not available. We also observed several misuses of SacreBLEU, such as the comparison of scores obtained by SacreBLEU against scores obtained by other tools. SacreBLEU signatures are also often not reported despite being required to ensure the comparability between SacreBLEU scores.

Ultimately, comparisons with copied scores must be avoided. As we will show in the next section, copying scores also calls for more pitfalls.

\subsection{The Data Approximation}
In MT, datasets are mostly monolingual or parallel texts used in three different steps of an experiment: training a translation model, tuning/validating the model, and evaluating it. Henceforth, we denote these datasets as training, validating, and testing data, respective to these three steps. How these datasets are pre-processed strongly influences translation quality. MT papers regularly propose new methods or algorithms that aim at better exploiting training and/or validating data. Following the scientific method, we can then define these new methods/algorithms and datasets as independent variables of an MT experiment while the translation quality, approximated by metric scores, would be our dependent variable that we want to measure. Testing the impact of a new algorithm on our dependent variable requires to keep all other independent variables, such as datasets, unchanged. In other words, changing datasets (even slightly) and methods/algorithms in the same experiment cannot answer whether the change in metric scores is due to the datasets, methods/algorithms, or the combination of both. 

Relying on our annotation for A6, we examined how often MT papers compared MT systems for which the datasets and/or their pre-processing\footnote{For pre-processing, we checked, for instance, tokenization (framework and parameters), casing, subword segmentations (method and vocabulary size), data filtering, etc.} described in the papers are not exactly identical. Note that we only performed this comparison for papers that copied and compared metric scores from previous work. 
Here, we also excluded comparisons between systems performed to specifically evaluate the impact of new datasets, pre-processing methods, and human intervention or feedback (e.g., post-editing and interactive MT). If we had any doubt whether a paper belongs or not to this category, we excluded it. Consequently, our estimation can be considered as the lower bound.

\begin{table}[t]
\centering
\scriptsize
\begin{tabular}{p{72mm}}
\toprule
\texttt{--type transformer --mini-batch-fit   --valid-freq 5000 --save-freq 5000 --workspace 10000 --disp-freq 500  --beam-size 12 --normalize 1 --valid-mini-batch 16 --overwrite  --early-stopping 5  --cost-type ce-mean-words  --valid-metrics bleu --keep-best   --enc-depth 6 --dec-depth 6   --transformer-dropout 0.1  --learn-rate 0.0003   --lr-warmup 16000   --lr-decay-inv-sqrt 16000 --lr-report   --label-smoothing 0.1  --devices 0 1 2 3 4 5 6 7  --optimizer-params 0.9 0.98 1e-09 --clip-norm 5 --sync-sgd  --exponential-smoothing --seed 1234}\\
\bottomrule
\end{tabular}
\caption{\label{tab:marianparam} Hyper-parameters of Marian used for training our NMT systems.}
\end{table}

To illustrate the impact of modifications of these datasets on metric scores, we conducted experiments using the training, validating, and testing data of the WMT20 news translation tasks. We trained neural MT (NMT) systems with Marian\footnote{Version: v1.7.6 1d4ba73 2019-05-11 17:16:31 +0100} \citep{junczys-dowmunt-etal-2018-marian}, using the hyper-parameters in Table \ref{tab:marianparam}, on all the provided parallel data (``all'' configurations) and removed sentence pairs based on their length (``Max Len.''). This simple filtering step is usually applied for a more efficient training or due to some limits of the framework, method, or algorithm used. Yet, it is so common as a pre-processing step that it is rarely described in papers. As shown in Table~\ref{tab:diffdata}, we observed that BLEU scores vary by several points depending on the maximum length used for filtering. Another common pre-processing step is the truecasing of the datasets. While it is rather commonly performed by participants in the WMT translation shared tasks, how casing is handled is rarely mentioned in research papers. In our experiments, applying this step changed BLEU scores by more than 0.5 points. Further experiments applying language identification filtering or removing one corpus from the training data also lead to variations in metric scores. The best configurations according to metric scores do not use truecasing and has a maximum sentence length set at 120 (second row). A comparison of this configuration with the third row, which uses truecasing and a different maximum sentence length, cannot lead to the conclusion that truecasing decreases translation quality, since we changed two variables at the same time.

\begin{table}[t]
    \centering
    \scriptsize
    \begin{tabular}{lcccccc}
    \toprule
        \multirow{2}{*}{Data}&Max  & \multirow{2}{*}{tc} & \multicolumn{2}{c}{En$\rightarrow$De}  & \multicolumn{2}{c}{Ja$\rightarrow$En}   \\
        & Len. & & BLEU  & chrF & BLEU & chrF   \\ 
        \midrule 
        all & 120 & $\checkmark$ &30.9 & 0.599 &20.4 & 0.478 \\
        \midrule
        all & 120 &   & 31.5$^\spadesuit$ & 0.604$^\spadesuit$ & 21.1$^\spadesuit$ & 0.481 \\
        all & 100 & $\checkmark$  & 30.8 & 0.597 &20.5 & 0.476 \\
        all & 80 & $\checkmark$  & 29.8$^\blacklozenge$ & 0.584$^\blacklozenge$ & 20.0 & 0.471$^\blacklozenge$\\
        all & 60 & $\checkmark$ & 26.6$^\blacklozenge$ &0.549$^\blacklozenge$ & 18.5$^\blacklozenge$ & 0.453$^\blacklozenge$ \\
        \midrule
         lid filtered & 120 & $\checkmark$ & 30.3$^\blacklozenge$ & 0.596 & 20.7 & 0.480 \\
          \midrule
         - 1 corpus & 120 & $\checkmark$ & 30.7 & 0.600 &19.6$^\blacklozenge$ & 0.468$^\blacklozenge$ \\
    
    \bottomrule    
    \end{tabular}
    \caption{BLEU and chrF scores of systems using differently pre-processed parallel data for training. ``Max Len.'' denotes that sentence pairs with sentence longer than the specified number, in terms of subword tokens, are removed. ``tc'' denotes whether truecasing is done or not. If not, original case of the data is kept for all datasets. ``lid filtered'' denotes that the training data are filtered with language identification tools. Last row denotes that we remove one corpus from the training data: ``Rapid'' for En$\rightarrow$De and ``OpenSubtitles'' for Ja$\rightarrow$En. $^\blacklozenge$ and $^\spadesuit$ respectively denote systems that are significantly worse and better ($p$-value $<$ 0.05), according to the metric, than the system in the first row.}
    \label{tab:diffdata}
\end{table}

\begin{figure}[t]
   \centering
   \includegraphics[scale=0.51]{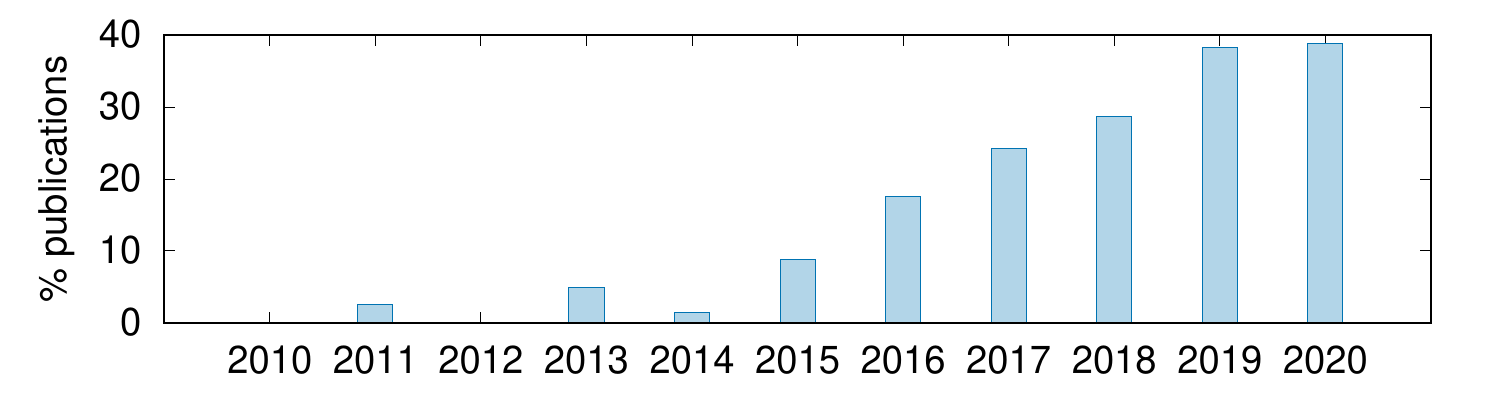}
   \caption{\label{fig:datadiffpaper} Percentage of papers that compared MT systems using data that are not identical.}
\end{figure}
While these observations may be expected or even obvious, Figure~\ref{fig:datadiffpaper} shows that we found in our meta-evaluation an increasing amount of MT papers (38.5\% for the 2019--2020 period) drawing conclusions of the superiority of a particular method or algorithm while also using different data.
While their conclusions may be valid, the evaluation conducted in these papers is scientifically flawed and cannot support the conclusions. We assume that this is mainly due to a rather common lack of detailed experimental settings. Consequently, it makes a specific experiment often impossible to be reproduced identically. In most cases, ensuring the comparability with the published scores of an MT system is only possible by replicating the MT system by ourselves. There have been initiatives towards the release of pre-processed datasets for MT, for instance by the WMT conference that released pre-processed data for WMT19.\footnote{This effort has not been conducted for WMT20.} Nonetheless, we only noticed a very small number of papers exploiting pre-processed training/validating/testing data publicly released by previous work.\footnote{For instance, \citet{ma-etal-2020-simple} and \citet{kang-etal-2020-dynamic} used exactly the same pre-processed data for research on document-level NMT released by \citet{maruf-etal-2019-selective}.}
We believe that the current trend should be reversed. Reviewers should also request more rigor to the authors by checking the configurations of the compared MT systems to make sure that their comparison can, indeed, answer whether the proposed method/algorithm improves MT independently of the data and their pre-processing.

\begin{figure}[t]
   \centering
   \includegraphics[scale=0.51]{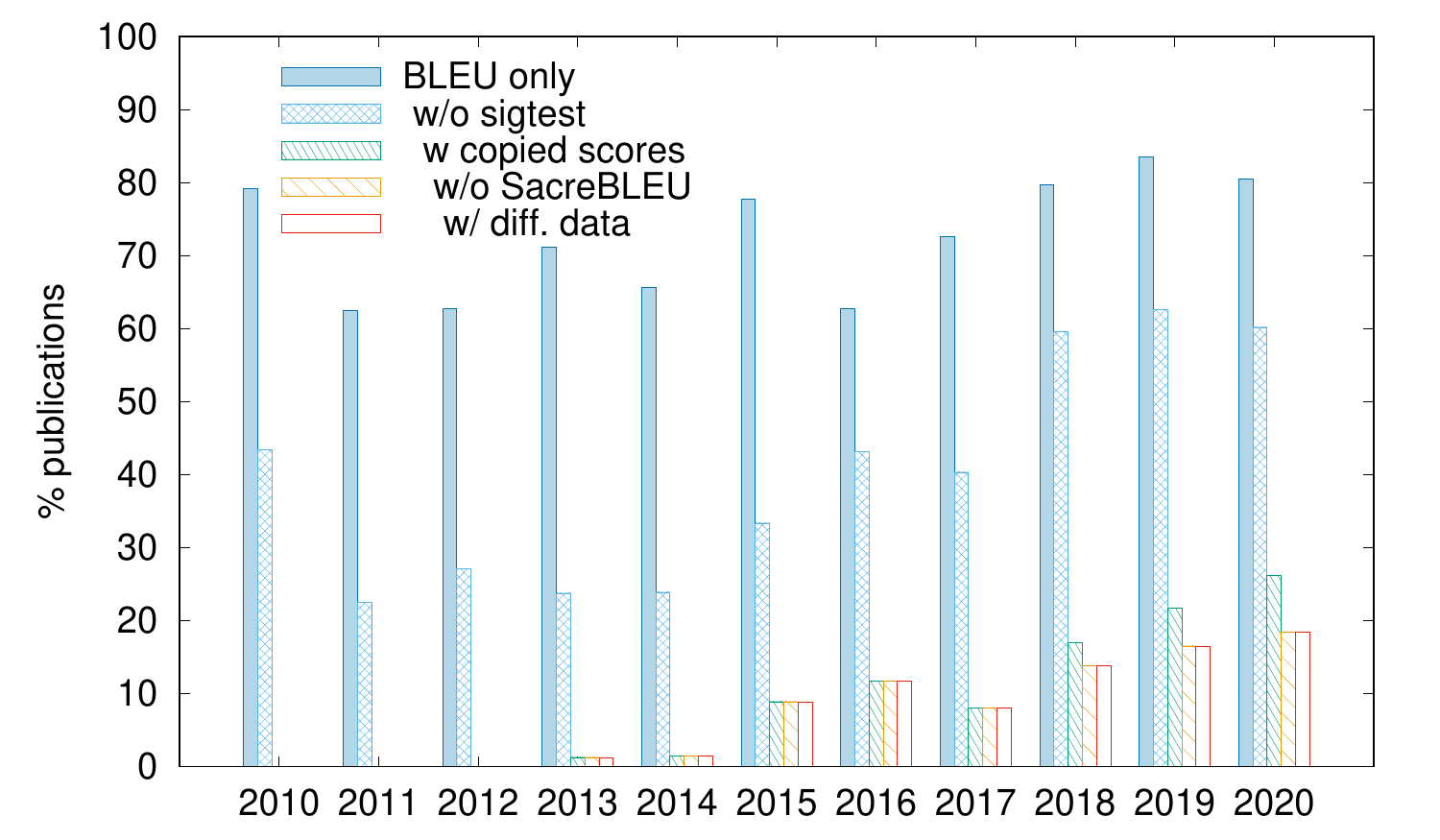}
   \caption{\label{fig:accumulation} Percentage of papers affected by the accumulation of pitfalls. Each bar considers only the papers counted by the previous bar, e.g., the last bar considers only papers that compared MT systems exploiting different datasets while exclusively using BLEU (``BLEU only''), without performing statistical significance testing (``w/o sigtest''), to measure differences with BLEU scores copied from other papers (``w/ copied scores'') while not using SacreBLEU (``w/o SacreBLEU'').}
\end{figure}

\section{A Guideline for MT Meta-Evaluation}
\label{sec:guideline}
\subsection{Motivation}
The MT community is well-aware of all the pitfalls described in Section~\ref{sec:pitfalls}. They have all been described by previous work. Nonetheless, our meta-evaluation shows that most MT publications are affected by at least one of these pitfalls. More puzzling are the trends we observed. Figure~\ref{fig:accumulation} shows that an increasing number of publications accumulate questionable evaluation practices. In the period 2019--2020, 17.4\% (38 papers) of the annotated papers exclusively relied for their evaluation on differences between BLEU scores of MT systems, of which at least some have been copied from different papers, without using SacreBLEU nor statistical significance testing, while exploiting different datasets.

While these pitfalls are known and relatively easy to avoid, they are increasingly ignored and accumulated. We believe that a clear, simple, and well-promoted guideline must be defined for automatic MT evaluation. Such a guideline would be useful only if it is adopted by authors and its application is checked by reviewers. For the latter, we also propose a simple scoring method for the meta-evaluation of MT.

Note that the proposed guideline and scoring method only cover the aspects discussed in this paper. Thus, their strict adherence can only guarantee a better evaluation but not a flawless evaluation.
\subsection{The Guideline}
This guideline and the scoring method that follows are proposed for MT papers that rely on automatic metric scores for evaluating translation quality.

\begin{enumerate}
    \item An MT evaluation may not exclusively rely on BLEU. Other automatic metrics that better correlate with human judgments, or a human evaluation, may be used in addition or in lieu of BLEU.
    \item Statistical significance testing may be performed on automatic metric scores to ensure that the difference between two scores, whatever its amplitude, is not coincidental.
    \item Automatic metric scores copied from previous work may not be compared. If inevitable, copied scores may only be compared with scores computed in exactly the same way, through tools guaranteeing this comparability, while providing all the necessary information to reproduce them.  
    \item Comparisons between MT systems through their metric scores may be performed to demonstrate the superiority of a method or an algorithm only if the systems have been trained, validated, and tested with exactly the same pre-processed data, unless the proposed method or algorithm is indeed dependent on a particular dataset or pre-processing. 
\end{enumerate}

The purpose of the following scoring method is to assess the trustworthiness of an automatic evaluation performed in an MT paper. Ultimately, it can be used for authors' self-assessment or by MT program committees to identify trustworthy papers.

Each ``yes'' answer to the following questions brings 1 point to the paper for a maximum of 4 points.
\begin{enumerate}
    \item Is a metric that better correlates with human judgment than BLEU used or is a human evaluation performed? 
    \item Is statistical significance testing performed?
    \item Are the automatic metric scores computed for the paper and not copied from other work?
        \\If copied, are all the copied and compared scores computed through tools that guarantee their comparability (e.g., SacreBLEU)?
    \item If comparisons between MT systems are performed to demonstrate the superiority of a method or an algorithm that is independent from the datasets exploited and their pre-processing, are all the compared MT systems exploiting exactly the same pre-processed data for training, validating, and testing? (if not applicable, give 1 point by default)
\end{enumerate}

\begin{figure}[t]
   \centering
   \includegraphics[scale=0.51]{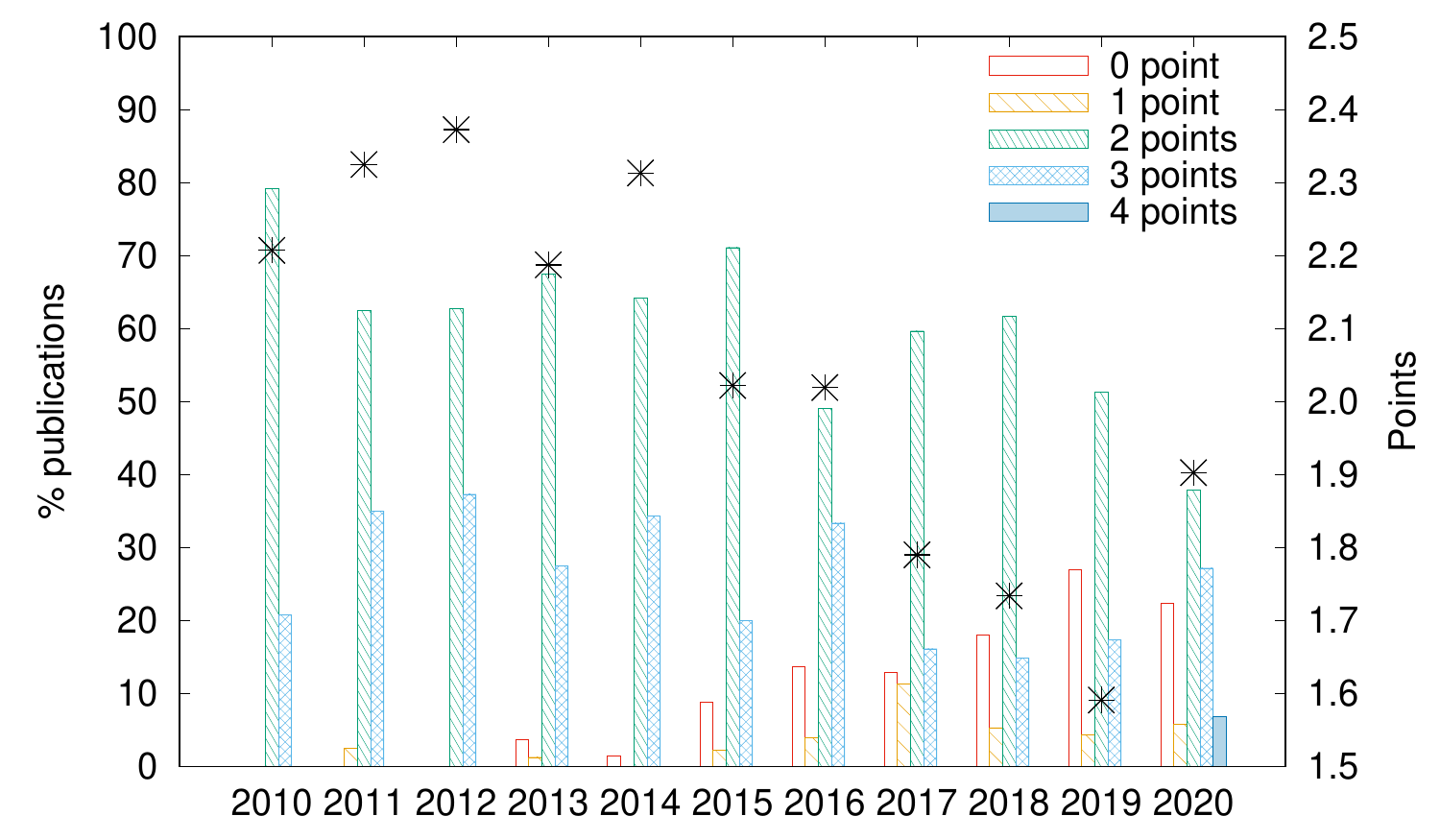}
   \caption{\label{fig:points} Average meta-evaluation score per year.}
\end{figure}

We scored all the annotated papers, and report on the average score and score distribution for each year in Figure~\ref{fig:points}. Based on this meta-evaluation, MT evaluation worsens.

\section{Conclusion}
Our meta-evaluation identified pitfalls in the MT evaluation in most of the annotated papers. The accumulation of these pitfalls and the concerning trends we observed lead us to propose a guideline for automatic MT evaluation. We hope this guideline, or a similar one, will be adopted by the MT community to enhance the scientific credibility of MT research.

This work also has its limitations since it does not cover all the pitfalls of MT evaluation. For instance, we noticed that MT papers regularly rely on the same language pairs to claim general improvements of MT. They also almost exclusively focus on translation from or into English. Another, more positive observation, is that MT papers tend to use stronger baseline systems, following some of the recommendations by \citet{denkowski-neubig-2017-stronger}, than at the beginning of the last decade when baseline systems were mostly vanilla MT systems.
For future work, we plan to extend our meta-evaluation of MT to publications at conferences in other research domains, such as Machine Learning and Artificial Intelligence.

As a final note, we would like to encourage NLP researchers to perform a similar meta-evaluation in their respective area of expertise. As we showed, it can unveil pitfalls and concerning trends that can be reversed before becoming prevalent.

\section*{Acknowledgments}
We would like to thank the reviewers for their insightful comments and suggestions. This work was partly supported by JSPS KAKENHI grant numbers 20K19879 and 19H05660.
\bibliography{acl2021}
\bibliographystyle{acl_natbib}

\appendix

\end{document}